\documentclass[10pt,twocolumn,letterpaper]{article}

\usepackage{cvpr}              %
\usepackage[accsupp]{axessibility}  %

\definecolor{cvprblue}{rgb}{0.21,0.49,0.74}
\usepackage[pagebackref,breaklinks,colorlinks,allcolors=cvprblue]{hyperref}

\usepackage{soul}
\usepackage{tikz}
\usepackage{bm}
\usepackage{algorithmicx}
\usepackage{algorithm}
\usepackage{algpseudocode}
\usepackage[accsupp]{axessibility}
\usepackage{pifont}
\usepackage{wasysym}
\usepackage{placeins} %
\usepackage{graphicx}
\usepackage[title]{appendix} 
\usepackage{caption} %
\usepackage{lipsum} %
\usepackage{subcaption} %
\usepackage{multirow}
\usepackage[hang,flushmargin]{footmisc}
\usepackage{balance}

\newcommand{\proj}[0]{\text{GaussianSpa}}

\def\paperID{7565} %
\def\confName{CVPR}
\def\confYear{2025}

\title{\proj: An ``Optimizing-Sparsifying" Simplification Framework for Compact and High-Quality 3D Gaussian Splatting}

\author{Yangming Zhang*~~~~Wenqi Jia*\\
Dept. of Computer Science\\
University of Texas at Arlington\\
{\tt\small \{yxz0925, wxj1489\}@mavs.uta.edu}
\and
Wei Niu\\
School of Computing\\
University of Georgia\\
{\tt\small wniu@uga.edu}
\and
Miao Yin$^\dagger$\\
Dept. of Computer Science\\
University of Texas at Arlington\\
{\tt\small miao.yin@uta.edu}
}

\begin{document}
\twocolumn[{
\maketitle
\centering
    \captionsetup{type=figure}
    \includegraphics[width=1\textwidth]{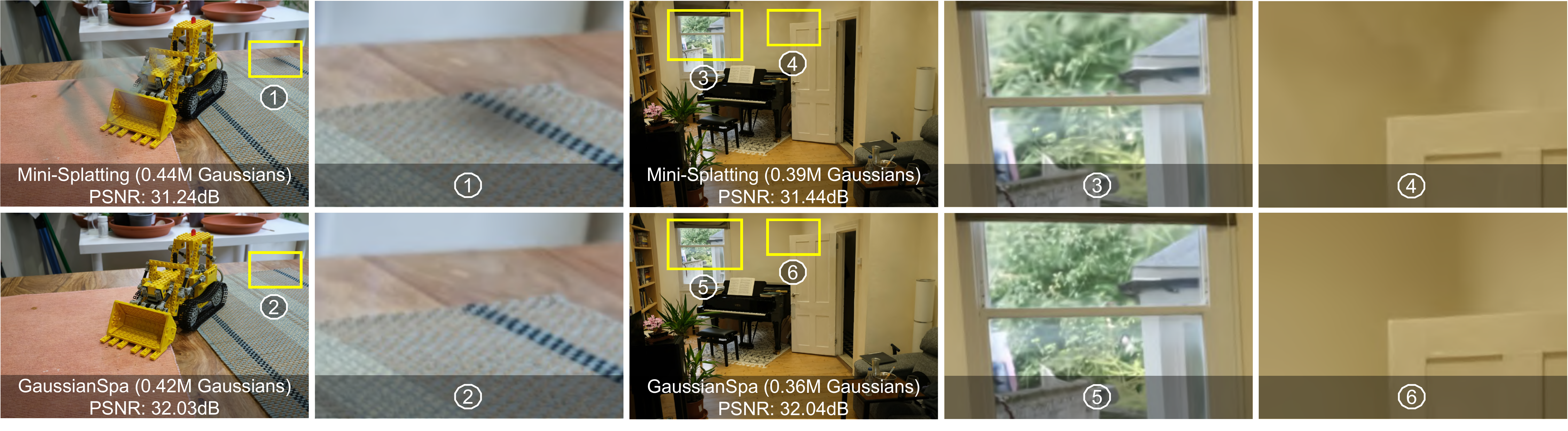}
    \vspace{-8.2mm}
    \captionof{figure}{\textbf{We present \proj, enabling high-quality and compact view synthesis with superior rendering of details.} Compared to the existing state-of-the-art method, Mini-Splatting \cite{fang2024mini}, our \textbf{GaussianSpa} captures detail-rich textures and low-frequency background more accurately with fewer Gaussians.}
    \label{fig:intro-cover-comp}
    \vspace{4.2mm}
}]
\begin{abstract}
\let\thefootnote\relax\footnotetext{*Yangming Zhang is the leading co-first author, while Wenqi Jia is the secondary co-first author. $^\dagger$Miao Yin is the corresponding author.}3D Gaussian Splatting (3DGS) has emerged as a mainstream for novel view synthesis, leveraging continuous aggregations of Gaussian functions to model scene geometry. However, 3DGS suffers from substantial memory requirements to store the large amount of Gaussians, hindering its efficiency and practicality. To address this challenge, we introduce \textbf{GaussianSpa}, an optimization-based simplification framework for compact and high-quality 3DGS. Specifically, we formulate the simplification objective as a constrained optimization problem associated with the 3DGS training. Correspondingly, we propose an efficient ``optimizing-sparsifying'' solution for the formulated problem, alternately solving two independent sub-problems and gradually imposing substantial sparsity onto the Gaussians in the 3DGS training process. We conduct quantitative and qualitative evaluations on various datasets, demonstrating the superiority of \proj~over existing state-of-the-art approaches. Notably, \proj~achieves \textbf{an average PSNR improvement of 0.9 dB on the real-world Deep Blending dataset with 10$\times$ fewer Gaussians} compared to the vanilla 3DGS. Our project page is available at \href{https://noodle-lab.github.io/gaussianspa}{https://noodle-lab.github.io/gaussianspa}.
\end{abstract}
    
\section{Introduction}
\label{sec:intro}

Novel view synthesis has become a pivotal area in computer vision and graphics, advancing applications such as virtual reality, augmented reality, and immersive media experiences \cite{gao2022nerf}. NeRF \cite{mildenhall2021nerf} has recently gained prominence in this domain because it can generate high-quality, photorealistic images from sparse input views by representing scenes as continuous volumetric functions based on neural networks. However, NeRF requires substantial computational resources and long training times, making it less practical for real-time and large-scale applications.

3D Gaussian Splatting (3DGS) \cite{kerbl20233d} has emerged as a powerful alternative, leveraging continuous aggregations of Gaussian functions to model scene geometry and appearance. Unlike NeRF, which relies on neural networks to approximate volumetric radiance fields, 3DGS directly represents scenes using a collection of Gaussians. This approach effectively captures details and smooth transitions, offering faster training and rendering. 3DGS achieves superior visual fidelity \cite{kopanas2021point} compared to NeRF while reducing computational overhead, making it more suitable for interactive applications that demand quality and performance.

Despite its strengths, 3DGS suffers from significant memory requirements that hinder its practicality. The main issue is the massive memory consumption required to store numerous Gaussians, each with parameters like position, covariance, and color. In densely sampled scenes, the sheer volume of Gaussians leads to memory usage that exceeds the capacity of typical hardware, making it challenging to handle higher-resolution scenes and limiting its applicability in resource-constrained environments.

\begin{figure}[t]
    \centering
    \includegraphics[width=0.9\linewidth]{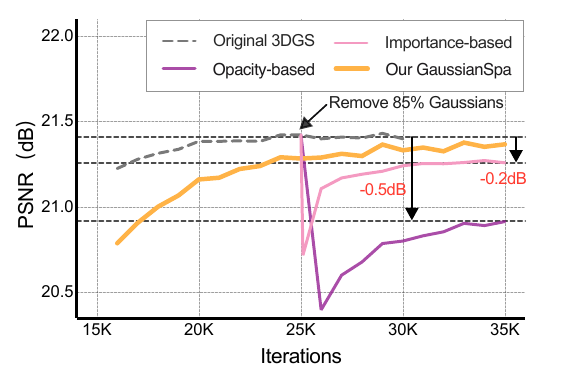}
    \vspace{-4mm}
    \caption{\textbf{PSNR curves of hand-crafted criteria-based pruning methods.} Gaussians are removed by 85\% at iteration 25K.}
    \vspace{-2mm}
    \label{fig:intro-sudden-psnr-drop}
\end{figure}

Existing works, e.g., Mini-Splatting \cite{fang2024mini}, LightGaussian \cite{fan2023lightgaussian}, LP-3DGS \cite{zhang2024lp}, EfficientGS \cite{liu2024efficientgs}, and RadSplat \cite{niemeyer2024radsplat}, have predominantly focused on mitigating this issue by removing a certain number of Gaussians. Techniques such as pruning and sampling aim to discard unimportant Gaussians based on hand-crafted criteria such as opacity \cite{kerbl20233d, yang2024spectrally}, importance score (hit count) \cite{fang2024mini, fan2023lightgaussian, niemeyer2024radsplat}, dominant primitives \cite{liu2024efficientgs}, and binary mask \cite{zhang2024lp, lee2024compact}. However, such single-perspective heuristic criteria may lack robustness in dynamic scenes or varying lighting. Moreover, the sudden one-shot removal may cause permanent loss of Gaussians crucial to visual synthesis, making it challenging to recover the original performance after even long-term training, as shown in Figure \ref{fig:intro-sudden-psnr-drop}. Mask-based pruning methods also suffer from this issue due to weak sparsity enforcement on the multitude of Gaussians. Consequently, while these methods can alleviate memory and storage burdens to some extent, they often lead to sub-optimal rendering outcomes with loss of details and visual artifacts, thereby compromising the quality of the synthesized views. 

In this paper, we present an optimization-based simplification framework, \proj, for compact and high-quality 3DGS. In the proposed framework, we formulate 3DGS simplification as a constrained optimization problem under a target number of Gaussians. Then, we propose an efficient ``optimizing-sparsifying" solution for the formulated problem by splitting it into two simple sub-problems that are alternately solved in the ``optimizing" step and the ``sparsifying" step. Instead of permanently removing a certain number of Gaussians, \proj~incorporates the ``optimizing-sparsifying'' algorithm into the training process, gradually imposing a substantial sparse property onto the trained Gaussians. Hence, our \proj~can simultaneously enjoy maximum information preservation from the original Gaussians and a desired number of reduced Gaussians, providing compact 3DGS models with high-quality rendering. Overall, our contributions can be summarized as follows:
\begin{itemize}
    \item We propose a general 3DGS simplification framework that formulates the simplification objective as an optimization problem and solves it in the 3DGS training process. In solving the formulated optimization problem, our proposed framework gradually restricts Gaussians to the target sparsity constraint without explicitly removing a specific number of points. Hence, \proj~can maximally maintain and smoothly transfer the information to the sparse Gaussians from the original model.
    
    \item We propose an efficient ``optimizing-sparsifying" solution for the formulated problem, which can be integrated into the 3DGS training with negligible costs, separately solving two sub-problems. In the ``optimizing" step, we optimize the original loss function attached by a regularization with gradient descent. In the ``sparsifying" step, we analytically project the auxiliary Gaussians onto the constrained sparse space.
    
    \item We comprehensively evaluate \proj~via extensive experiments on various complex scenes, demonstrating improved rendering quality compared to existing state-of-the-art approaches. Particularly, with as high as 10$\times$ fewer number of Gaussians than the vanilla 3DGS, \proj~achieves an average 0.4 dB PSNR improvement on the Mip-NeRF 360 \cite{barron2022mip} and Tanks\&Temples \cite{Knapitsch2017} datasets, 0.9 dB on Deep Blending \cite{hedman2018deep} dataset. Furthermore, we conduct various visual quality analyses, demonstrating the superiority of \proj~in rendering high-quality views and capturing rich details. 
    
\end{itemize}

\section{Related Work}
\label{sec:related}

\begin{figure*}[t!]
    \centering
    \includegraphics[width=\linewidth]{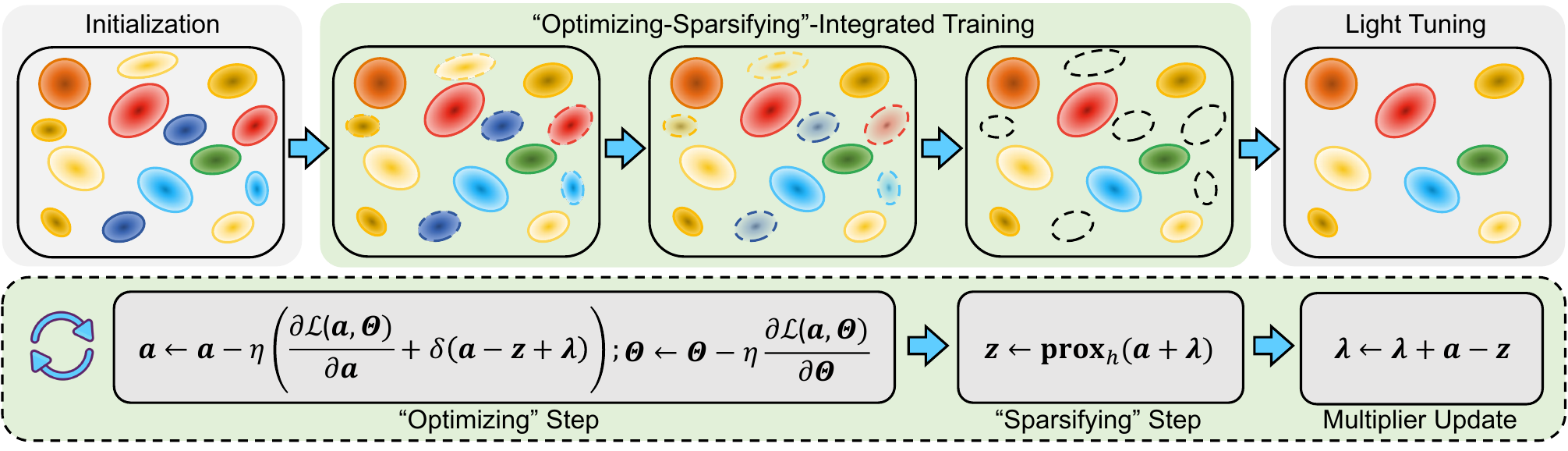}
    \vspace{-7mm}
    \caption{\textbf{Overall workflow of our proposed \proj~framework.}}
    \vspace{-2mm}
    \label{fig:overall-framework}
\end{figure*}

\subsection{Novel View Synthesis}
Novel View Synthesis (NVS) \cite{avidan1997novel} generates images of a 3D scene from unseen viewpoints based on existing multi-view data. A seminal work in this field is NeRF \cite{mildenhall2021nerf}, which uses an implicit neural network to represent a scene as a continuous 5-dimensional function, mapping 3D coordinates and viewing directions to color and density. However, its reliance on single narrow-ray sampling poses challenges in addressing aliasing and blur artifacts. More recently, 3D Gaussian Splatting (3DGS), which leverages continuous aggregations of Gaussian functions to model scene geometry, has emerged as a point-based alternative, demonstrating significant improvements in rendering quality and speed.

\subsection{Efficient 3DGS}

3DGS significantly improves rendering quality and computational efficiency compared to traditional methods such as NeRF \cite{mildenhall2021nerf}, which are often resource-intensive and computationally slow. However, the demands of real-time rendering and the reconstruction of unbounded, large-scale 3D scenes underscore the requirement for model efficiency \cite{chen2024survey}. Optimization for efficient 3DGS focuses on four main categories: Gaussian simplification, quantization, spherical harmonics optimization, and hybrid compression.

\textbf{Gaussian Simplification.}
Gaussian simplification involves removing Gaussians to compress 3DGS models into compact forms, significantly reducing memory costs. Prior approaches \cite{fang2024mini, niemeyer2024radsplat, mallick2024taming, zhang2024lp, lu2024scaffold, ren2024octree, liu2024atomgs, cheng2024gaussianpro, kim2024color} mainly focus on pruning redundant Gaussians according to hand-crafted importance criteria.
Mini-Splatting \cite{fang2024mini} addresses overlapping and reconstruction issues with blur splitting, depth reinitialization, and stochastic sampling. Radsplatting \cite{niemeyer2024radsplat} enhances robustness by using a max operator to compute importance scores from ray contributions. Taming 3DGS \cite{mallick2024taming} applies selective densification based on pixel saliency and gradients, while LP-3DGS \cite{zhang2024lp} learns a binary mask for efficient pruning.

\textbf{Quantization.}
Quantization \cite{van2017neural} aims to compress data using discrete entries without appreciable degradation in quality, reducing data redundancy. Several works \cite{girish2023eagles, navaneet2024compgs, chen2025hac, wang2024end, papantonakis2024reducing, morgenstern2023compact, niedermayr2024compressed} apply quantization for compressing 3DGS models.
For example, EAGLES \cite{girish2023eagles} quantizes attributes like spherical harmonics (SH) coefficients. 
Vector quantization, a form of quantization, compresses data by mapping vectors to a learned codebook. 
CompGS \cite{navaneet2024compgs} leverages K-means-based vector quantization to reduce storage and rendering time, while RDO-Gaussian \cite{wang2024end} combines pruning and entropy-constrained vector quantization for efficient compression with minimal quality loss.

\textbf{Spherical Harmonics Optimization.}
Among 3DGS parameters, spherical harmonics (SH) coefficients take up a significant portion, requiring (45+3) floating-point values per Gaussian for a degree of 3 \cite{kerbl20233d}, which accounts for around 80\% of the total attribute volume. Prior works \cite{fan2023lightgaussian, lee2024compact} optimize SH coefficients to reduce storage overhead. For instance, EfficientGS \cite{liu2024efficientgs} retains zeroth-order SH for all Gaussians, increasing the order when necessary.

\textbf{Hybrid Compression.}
In addition to methods focused on individual aspects of 3DGS compression, several recent approaches \cite{fan2023lightgaussian, liu2024efficientgs, lee2024compact, xie2024mesongs} combine multiple techniques for hybrid compression to achieve further storage reduction.
LightGaussian \cite{fan2023lightgaussian} uses a global importance score for pruning and applies vector quantization to compress SH coefficients. EfficientGS \cite{liu2024efficientgs} selectively increases SH order and densifies non-steady Gaussians. CompactGaussian \cite{lee2024compact} introduces a learnable mask for pruning, uses residual vector quantization for scale and rotation, and replaces SH with a grid-based neural field for color representation.

Although mask-based method impose sparsity relying on mask loss, they and other methods both lead to irreversible performance loss in the Gaussian simplification process by performing a one-shot pruning on the Gaussians, as shown in Figure \ref{fig:intro-sudden-psnr-drop}. Consequently, performance often worsens when applying SH optimization and quantization due to accumulated errors.
Our work focuses on \textit{Gaussian simplification} by formulating it as an optimization problem and imposing substantial sparsity onto Gaussians, mitigating performance issues with improved rendering quality.

\section{Methodology: \proj}
\label{sec:method}

\begin{figure*}[t!]
    \centering
    \includegraphics[width=0.98\linewidth]{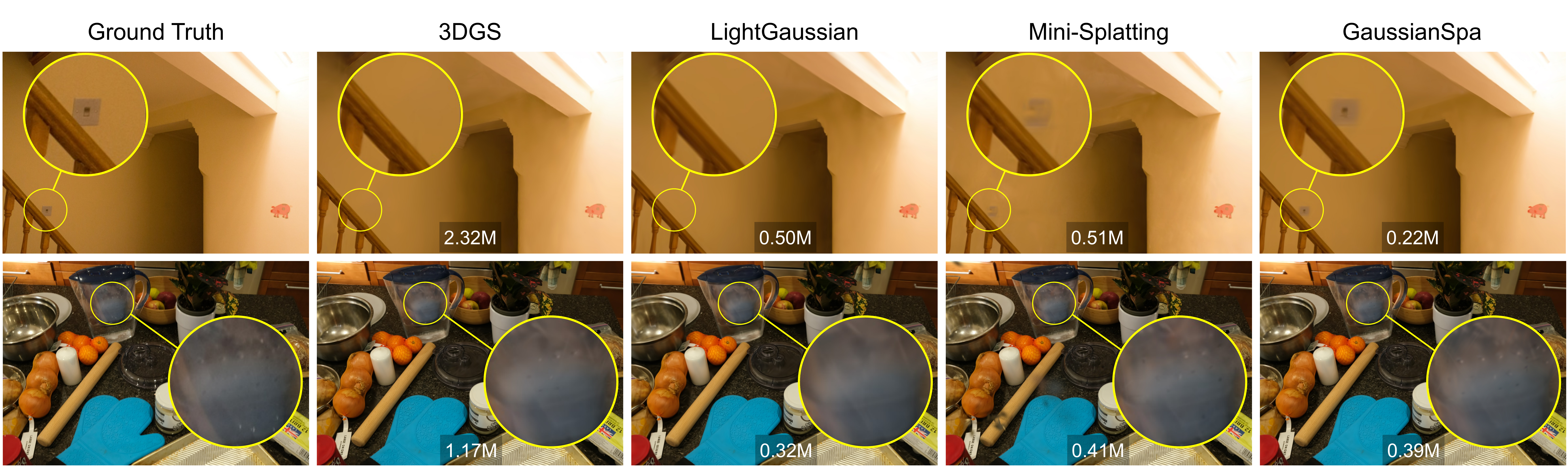}
    \vspace{-3mm}
    \caption{\textbf{Visual quality results on the Drjohnson and Counter scenes, compared to existing simplification approaches and vanilla 3DGS.} The numbers of remaining Gaussians are displayed. It is observed that our \proj~recovers details closest to the ground truth in the actual rendering outcomes with a significantly reduced number of Gaussians.}
    \vspace{-1mm}
    \label{fig:RGBrender4exp}
\end{figure*}

\subsection{Background of 3D Gaussian Splatting}
3D Gaussian Splatting (3DGS) explicitly represents scenes using a collection of point-based 3D continuous Gaussians. Specifically, each Gaussian $G$ is characterized by its covariance matrix $\bm{\Sigma}$ and center position $\bm{\mu}$ as
\begin{equation}\label{eqn:gaussian-defination}
  G(\bm{x})=\exp\left(-\frac{1}{2}(\bm{x}-\bm{\mu})^\top\bm{\Sigma}^{-1}(\bm{x}-\bm{\mu})\right),
\end{equation}
where $\bm{x}$ is an arbitrary location in the 3D scene. The covariance matrix $\bm{\Sigma}$ is generally composed of $\bm{\Sigma}=\bm{R}\bm{S}\bm{S}^\top\bm{R}^\top$ with a rotation matrix $\bm{R}$ and a scale matrix $\bm{S}$, ensuring the positive definite property. Each Gaussian also has an associated opacity $a$ and spherical harmonics (SH) coefficients.

When rendering a 2D image from the 3D scene at a certain camera angle, 3DGS projects 3D Gaussians to that 2D plane and calculates the projected 2D covariance matrix by
\begin{equation}\label{eqn:cov-projection}
  \bm{\Sigma'}=\bm{J}\bm{W}\bm{\Sigma}\bm{W}^\top\bm{J}^\top,
\end{equation}
where $\bm{W}$ is the view transformation, and $\bm{J}$ is the Jacobian of the affine approximation of the projective transformation \cite{964490}. Then, for each pixel on the 2D image, the color $C$ is computed by blending all $N$ ordered Gaussians contributing to the pixel as
\begin{equation}\label{eqn:render-color}
  C=\sum_{i\in N}c_i\alpha_i \prod_{j=1}^{i-1}(1-\alpha_j).
\end{equation}
\vspace{-0.5mm}
Here, $c_i$ and $\alpha_i$ denote the view-dependent color computed from the corresponding SH coefficients and the rendering opacity calculated by the Gaussian opacity $a_i$, respectively, corresponding to the $i$-th Gaussian.

In the training process \cite{kerbl20233d}, the Gaussians are initialized from a sparse point cloud generated by Structure-from-Motion (SfM) \cite{schonberger2016structure}. Then, the number of Gaussians is adjusted by a densification algorithm \cite{kerbl20233d} according to the reconstruction geometry. Afterward, each Gaussian's attributes, including the center position, rotation matrix, scaling matrix, opacity, and SH coefficients, are optimized by minimizing the reconstruction errors using the standard gradient descent. Specifically, the loss $\mathcal{L}$ is given by\vspace{-1mm}
\begin{equation}
  \mathcal{L}=(1-\rho)\mathcal{L}_1+\rho\mathcal{L}_{\text{D-SSIM}},
  \vspace{-1mm}
\end{equation}
where $\mathcal{L}_1$ and $\mathcal{L}_{\text{D-SSIM}}$ are the $\ell_1$ norm and the structural dissimilarity between the rendered output and the ground truth, respectively. $\rho$ is a parameter that controls the tradeoff between the two losses. 

\subsection{Problem Formulation: Simplification as Constrained Optimization}

To mitigate irreversible information loss caused by existing one-shot criterion-based pruning, our key innovation is to \ul{gradually impose substantial sparsity onto the Gaussians and maximally preserve information in the training process}. Mathematically, considering a general 3DGS training, we introduce a sparsity constraint on the Gaussians to the training objective, $\min\mathcal{L}$. Thus, the 3DGS training is reformulated as another constrained optimization problem, the primary formulation of our proposed simplification framework, i.e.,
\begin{equation}
\begin{aligned}\label{eqn:obj-original-form}
  \min ~&\mathcal{L},\\
  \text{s.t.}~~ &\mathcal{N}(\bm{G})\leq\kappa.
\end{aligned}
\vspace{-0.5mm}
\end{equation}
The introduced constraint $\mathcal{N}(\cdot)\le\kappa$ restricts the number of Gaussians no greater than a target number $\kappa$.

However, the above problem cannot be solved because the restriction on the number of Gaussian functions is not differentiable. Recalling the rendering process with Eq. \ref{eqn:render-color}, the contribution of the $i$-th Gaussian to a particular pixel is finally determined by its opacity $a_i$. Alternatively, the original sparsity constraint directly on the Gaussians can be explicitly transformed into a sparsity constraint on the opacity variables. As each Gaussian only has one opacity variable, we can use a vector $\bm{a}$ to represent all the opacities and easily restrict its number of non-zeros with $\ell_0$ norm, $\|\bm{a}\|_0$. For better presentation, we use a variable set $\bm{\Theta}$ to represent other GS variables except for the opacity $\bm{a}$. Hence, the constrained optimization problem can be reformulated as\vspace{-1mm}
\begin{equation}\label{eqn:obj-opacity-form}
\begin{aligned}
  \min_{\bm{a},\bm{\Theta}} ~&\mathcal{L}(\bm{a},\bm{\Theta}),\\
  \text{s.t.}~ ~& \|\bm{a}\|_0\leq\kappa.
\end{aligned}
\vspace{-1mm}
\end{equation}
With this reformulation, we have simplified the primary problem to another form that our proposed ``optimizing-sparsifying" can efficiently solve.

\subsection{``Optimizing-Sparsifying'' Solution}

\begin{algorithm}[t!]
\newcommand{\assign}{$\leftarrow$}
\caption{Procedure of ``Optimizing-Sparsifying"}\label{alg:optimizing-sparsifying}
\textbf{Input:} Gaussian opacity $\bm{a}$, 3DGS variables $\bm{\Theta}$, target number of Gaussians $\kappa$, penalty parameter $\delta$, feasibility tolerance $\epsilon$, maximum iterations $T$.\par
\textbf{Output:} Optimized $\bm{a}$ and $\bm{\Theta}$.
\begin{algorithmic}[1]
\State $\bm{z}$ \assign $\bm{a}$, $\bm{\lambda}$ \assign $\bm{0}$;
\State $t$ \assign $0$;
\While{$\|\bm{a} - \bm{z}\|^2 > \epsilon$ and $t \leq T$}
    \State Update $\bm{a}$ and $\bm{\Theta}$ with Eq. \ref{eqn:update-optimizing-step}; \Comment{\textcolor{Orange}{\textit{``Optimizing" Step}}}
    \State Update $\bm{z}$ with Eq. \ref{eqn:update-sparsifying-step};
    \Comment{\textcolor{Orange}{\textit{``Sparsifying" Step}}}
    \State Update $\bm{\lambda}$ with Eq. \ref{eqn:update-multiplier};
    \Comment{\textcolor{Orange}{\textit{Multiplier Update}}}
    \State $t$ \assign $t+1$;
\EndWhile
\end{algorithmic}
\end{algorithm}

In this subsection, we present an efficient ``optimizing-sparsifying" solution for problem Eq. \ref{eqn:obj-opacity-form}. The key idea is to split it into two simple sub-problems separately solved in an ``optimizing" step and a ``sparsifying" step alternately. 

In order to deal with the non-convex constraint, $\|\bm{a}\|_0\leq\kappa$, we first introduce an indicator function,\vspace{-1mm}
\begin{equation}\label{eqn:indicator-func}
  h(\bm{a})=\left\{\begin{matrix}
    0, & \|\bm{a}\|_0\le\kappa, \\
    +\infty, & \textup{otherwise}, \\
\end{matrix}\right.
\vspace{-1mm}
\end{equation}
to the minimization objective and remove the sparsity constraint. Thus, problem Eq. \ref{eqn:obj-opacity-form} becomes an unconstrained optimization form, i.e.,\vspace{-1mm}
\begin{equation}\label{eqn:obj-unconstrained}
  \min_{\bm{a},\bm{\Theta}}~\mathcal{L}(\bm{a},\bm{\Theta})+h(\bm{a}).
\vspace{-1mm}
\end{equation}
Here, the first term is the original 3DGS training objective, which is easy to handle, while the second term is still non-differentiable. This prevents the above problem from being solved using gradient-based solutions under existing training frameworks such as PyTorch \cite{paszke2019pytorch}. To decouple the two terms, we further introduce an auxiliary variable $\bm{z}$, whose size is identical to $\bm{a}$, and convert Eq. \ref{eqn:obj-unconstrained} as\vspace{-1mm}
\begin{equation}\label{eqn:obj-final}
\begin{aligned}
  \min_{\bm{a},\bm{z},\bm{\Theta}} ~&\mathcal{L}(\bm{a},\bm{\Theta})+h(\bm{z}),\\
  \text{s.t.}~ ~& \bm{a}=\bm{z}.
\end{aligned}
\vspace{-1mm}
\end{equation}

The above problem becomes another constrained problem with an equality constraint. Thus, the corresponding dual augmented Lagrangian \cite{boyd2004convex} form can be given by\vspace{-1mm}
\begin{equation}\label{eqn:obj-lagrangian}
  L(\bm{a},\bm{z},\bm{\Theta},\bm{\lambda};\delta)=\mathcal{L}(\bm{a},\bm{\Theta})+h(\bm{z})+\frac{\delta}{2}\|\bm{a}-\bm{z}+\bm{\lambda}\|^2+\frac{\delta}{2}\|\bm{\lambda}\|^2,
\end{equation}
where $\bm{\lambda}$ is the dual Lagrangian multiplier, and $\delta$ is the penalty parameter. Correspondingly, we can separately optimize over the 3DGS variables, $\bm{a},\bm{\Theta}$, and the auxiliary variable, $\bm{z}$, in the augmented Lagrangian, then solve two split sub-problems in an ``optimizing" step and a ``sparsifying" step alternately.\vspace{-1mm}
\textcolor{Blue}{
\begin{equation}\label{eqn:obj-optimizing}
  \hspace{-1.6mm}\RHD~\textbf{``Optimizing" Step:~~}\min_{\bm{a},\bm{\Theta}}\mathcal{L}(\bm{a},\bm{\Theta})+\frac{\delta}{2}\|\bm{a}-\bm{z}+\bm{\lambda}\|^2.
\end{equation}}

\begin{figure}[t!]
    \centering
    \includegraphics[width=\linewidth]{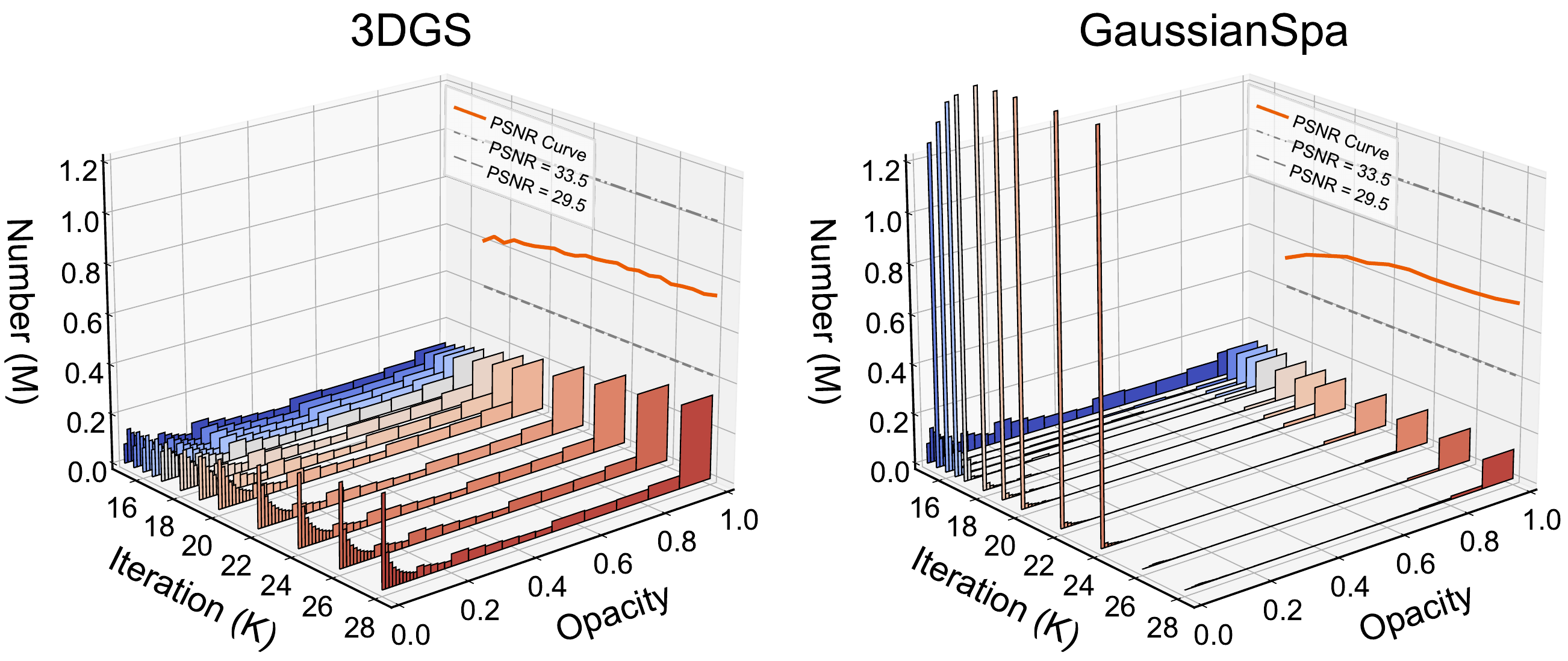}
    \vspace{-6mm}
    \caption{\textbf{Evolution of opacity distribution and PSNR for 3DGS and \proj.} The ``optimizing-sparsifying" starts at iteration 15K. At iteration 25K, \proj~removes ``zero" Gaussians, followed by a light tuning.}
    \vspace{-0.3mm}
\label{fig:opacity-distribution}
\end{figure}

\vspace{-4mm}In the ``optimizing" step, the minimization objective, Eq. \ref{eqn:obj-optimizing}, targets only 3DGS variables and contains two terms, which are the original 3DGS loss function and a quadratic regularization that enforces the opacity $\bm{a}$ close to the exactly sparse variable $\bm{z}$. Since both terms are differentiable, the sub-problem in this step can be directly optimized with gradient descent. In other words, the solution for this sub-problem is consistent with \textit{\textbf{optimizing}} 3DGS variables with additional gradients, which are computed by
\begin{align}
  \frac{\partial L}{\partial \bm{a}}&=\frac{\partial \mathcal{L}(\bm{a},\bm{\Theta})}{\partial \bm{a}}+\delta(\bm{a}-\bm{z}+\bm{\lambda}),\\
  \frac{\partial L}{\partial \bm{\Theta}}&=\frac{\partial \mathcal{L}(\bm{a},\bm{\Theta})}{\partial \bm{\Theta}}.
\end{align}
Hence, the opacity $\bm{a}$ and other GS variables $\bm{\Theta}$, can be updated by\vspace{-1mm}
\begin{equation}\label{eqn:update-optimizing-step}
\bm{a}\leftarrow\bm{a}-\eta \frac{\partial L}{\partial \bm{a}},~~
  \bm{\Theta}\leftarrow\bm{\Theta}-\eta \frac{\partial L}{\partial \bm{\Theta}},
\end{equation}
respectively. Here, $\eta$ is the learning rate during training.

\begin{table*}[h]
  \centering
    \resizebox{\linewidth}{!}{%
    \begin{tabular}{lcccc|cccc|cccc}
    \toprule
    \multirow{2}[2]{*}{Method} & \multicolumn{4}{c|}{Mip-NeRF 360} & \multicolumn{4}{c|}{Tanks\&Temples} & \multicolumn{4}{c}{Deep Blending} \\
\cmidrule{2-13}          & PSNR↑  & SSIM↑  & LPIPS↓ & \#G/M↓ & PSNR↑  & SSIM↑  & LPIPS↓ & \#G/M↓ & PSNR↑  & SSIM↑  & LPIPS↓ & \#G/M↓ \\
    \midrule
    3DGS \cite{kerbl20233d}  & 27.45 & 0.810 & 0.220 & 3.110 & 23.63 & 0.850 & 0.180 & 1.830 & 29.42 & 0.900 & 0.250 & 2.780 \\
    \midrule
    CompactGaussian \cite{lee2024compact}  & 27.08 & 0.798 & 0.247 & 1.388 & 23.32 & 0.831 & 0.201 & 0.836 & 29.79 & 0.901 & 0.258 & 1.060 \\
    LP-3DGS-R \cite{zhang2024lp} & 27.47 & 0.812 & 0.227 & 1.959 & 23.60 & 0.842 & 0.188 & 1.244 & -     & -     & -     & - \\
    LP-3DGS-M \cite{zhang2024lp} & 27.12 & 0.805 & 0.239 & 1.866 & 23.41 & 0.834 & 0.198 & 1.116 & -     & -     & -     & - \\
    EAGLES \cite{girish2023eagles} & 27.23 & 0.809 & 0.238 & 1.330 & 23.37 & 0.840 & 0.200 & 0.650 & 29.86 & 0.910 & 0.250 & 1.190 \\
    Mini-Splatting \cite{fang2024mini} & 27.40 & 0.821 & 0.219 & 0.559 & 23.45 & 0.841 & 0.186 & 0.319 & 30.05 & 0.909 & 0.254 & 0.397 \\
    Taming 3DGS \cite{mallick2024taming} & 27.31 & 0.801 & 0.252 & 0.630 & 23.95 & 0.837 & 0.201 & 0.290 & 29.82 & 0.904 & 0.260 & 0.270 \\
    CompGS \cite{navaneet2024compgs} & 27.12 & 0.806 & 0.240 & 0.845 & 23.44 & 0.838 & 0.198 & 0.520 & 29.90 & 0.907 & 0.251 & 0.550 \\
    \midrule
    \rowcolor[rgb]{ .902,  .902,  .902} &  &  &  &  &  &  &  &  & \textbf{30.37} & \textbf{0.914} & \textbf{0.249} & 0.335 \\
    \rowcolor[rgb]{ .902,  .902,  .902} \multirow{-2}{*}{\textbf{GaussianSpa}} & \multirow{-2}{*}{\textbf{27.85}} & \multirow{-2}{*}{\textbf{0.825}} & \multirow{-2}{*}{\textbf{0.214}} & \multirow{-2}{*}{\textbf{0.547}} & \multirow{-2}{*}{\textbf{23.98}} & \multirow{-2}{*}{\textbf{0.852}} & \multirow{-2}{*}{\textbf{0.180}} & \multirow{-2}{*}{\textbf{0.269}} & 30.33 & 0.912 & 0.254 & \textbf{0.256} \\
    \bottomrule
    \end{tabular}%
    }
    \vspace{-3mm}
    \caption{\textbf{Quantitative results on multiple datasets, compared with existing state-of-the-art works.} Mini-Splatting \cite{fang2024mini} results are replicated using official code. 3DGS results are reported from \cite{girish2023eagles}. ``\#G/M'' represents number of million Gaussians.
    }
    \vspace{-1mm}
  \label{tab:Quantitative_comparision}%
\end{table*}%

\vspace{-4mm}
\textcolor{Blue}{
\begin{equation}\label{eqn:obj-sparsifying}
  \hspace{-1.8mm}\RHD~\textbf{``Sparsifying" Step:~~}\min_{\bm{z}}h(\bm{z})+\frac{\delta}{2}\|\bm{a}-\bm{z}+\bm{\lambda}\|^2.
\end{equation}}

\vspace{-2mm}In the ``sparsifying" step, the minimization objective, Eq. \ref{eqn:obj-sparsifying}, targets only the auxiliary variable, $\bm{z}$, while the term of indicator function dominates Eq. \ref{eqn:obj-sparsifying}. This sub-problem essentially has a form of the proximal operator \cite{parikh2014proximal}, $\mathrm{\bf{prox}}_h$, associated with the indicator function $h(\cdot)$. Thus, the solution of this sub-problem can be directly given by 
\begin{equation}\label{eqn:update-sparsifying-step}
  \bm{z}\leftarrow\mathrm{\bf{prox}}_h(\bm{a}+\bm{\lambda}).
\end{equation}
The proximal operator maps the opacity $\bm{a}$ to the auxiliary variable $\bm{z}$, which contains at most $\kappa$ non-zeros. This operator results in a \textit{projection} solution \cite{boyd2011distributed}, \textit{\textbf{sparsifying}} $(\bm{a}+\bm{\lambda})$, by projecting it onto a set that satisfies $\|\bm{z}\|_0\le\kappa$. More specifically, the projection keeps top-$\kappa$ elements and sets the rest to zeros.

\noindent\textcolor{Blue}{$\RHD$~\textbf{Overall Procedure.}} After the ``optimizing" step and the ``sparsifying" step, the dual Lagrangian multiplier is updated by
\begin{equation}\label{eqn:update-multiplier}
  \bm{\lambda}\leftarrow\bm{\lambda}+\bm{a}-\bm{z}.
\end{equation}
Overall, the ``optimizing," ``sparsifying," and multiplier update steps are performed alternately in the 3DGS training process until convergence, as summarized in Algorithm \ref{alg:optimizing-sparsifying}. Generally, we consider our solution convergences once the condition, $\|\bm{a}-\bm{z}\|^2\le \epsilon$, is satisfied, or the iterations reach the predefined maximum number.

\textbf{Remark.} 
The ``optimizing" step optimizes over the original Gaussian variables and pushes the Gaussian opacity $\bm{a}$ close to the exactly sparse auxiliary variable $\bm{z}$ using a gradient-based approach, simultaneously satisfying the 3DGS performance and the sparsity requirements. The ``sparsifying" step essentially prunes the auxiliary variable $\bm{z}$, which can be considered the exactly sparse version of $\bm{a}$. As those steps are performed alternately, summarized in Algorithm \ref{alg:optimizing-sparsifying}, the process eventually converges to a sparse set that maximizes the performance. Upon the convergence, most Gaussians become ``zeros," and the information can be preserved and transferred to the ``non-zero" Gaussians.

\subsection{Overall Workflow of \proj}

With the proposed ``optimizing-sparsifying"-integrated training process presented in the previous subsection, the Gaussians exhibit substantial sparsity characteristics. In other words, a certain number of Gaussian opacities are close to zeros, meaning those Gaussians have almost no contribution to the rendering, which can be directly removed. As the redundant Gaussians are appropriately removed, our simplified 3DGS model can exhibit superior performance after a light tuning, even slightly higher than the original 3DGS. The overall workflow of our \proj~is illustrated in Figure \ref{fig:overall-framework}. 

We visualize the evolution of Gaussian opacity distribution along with the PSNR, trained on the Room scene, as shown in Figure \ref{fig:opacity-distribution}. The opacity distribution of \proj~is significantly distinct from the original 3DGS -- there is a clear gap between the ``zero" Gaussians and the remaining Gaussians in \proj, while the PSNR keeps increasing. This means \proj~has successfully imposed a substantial sparsity property and preserved information to the ``non-zero" Gaussians. At iteration 25K, we remove all ``zero" Gaussians and perform a light tuning that further improves the performance, finally obtaining a compact model with high-quality rendering.

\section{Experiments}
\label{sec:exp}

\begin{figure*}[t!]
    \centering
    \includegraphics[width=\linewidth]{figs/ellipsoid_train_compressed.pdf}
    \vspace{-8mm}
    \caption{\textbf{Visualization for rendered Gaussian ellipsoids.} (Add'l Figure \ref{fig:app-visual-results-ellipsoids} in the Supplementary Material). \proj~renders high-frequency (e.g., air outlet of the train) and low-frequency (e.g., large-area sky) areas adaptively using appropriate numbers of Gaussians, respectively, more precisely matching the original textures, compared to the state-of-the-art Mini-Splatting \cite{fang2024mini} and the vanilla 3DGS. }
    \label{fig:visual-results-ellipsoids}
    \vspace{-1mm}
\end{figure*}

\subsection{Experimental Settings}

\textbf{Datasets, Baselines, and Metrics.} We evaluate the rendering performance on multiple real-world datasets, including Mip-NeRF 360 \cite{barron2022mipnerf360}, Tanks\&Temples \cite{Knapitsch2017}, and Deep Blending \cite{hedman2018deep}. On all datasets, we compare our proposed \proj~with vanilla 3DGS \cite{kerbl20233d} and existing state-of-the-art Gaussian simplification approaches, including CompactGaussian \cite{lee2024compact}, LP-3DGS \cite{zhang2024lp}, EAGLES \cite{girish2023eagles}, Mini-Spatting \cite{fang2024mini}, Taming 3DGS \cite{mallick2024taming}, and CompGS \cite{navaneet2024compgs}. We follow the standard practice to quantitatively evaluate the rendering performance, reporting results using the peak signal-to-noise ratio (PSNR), structural similarity (SSIM), and learned perceptual image patch similarity (LPIPS). In addition to quantitative results, we comprehensively evaluate \proj~with visual analyses such as ellipsoid view distribution and point cloud visualization against the existing state-of-the-art approaches.

\textbf{Implementation Details.} We conduct experiments under the same environment specified in the original 3DGS \cite{kerbl20233d} with the PyTorch framework. Our experimental server has two AMD EPYC 9254 CPUs and eight NVIDIA GTX 6000 Ada GPUs. We use the same starting checkpoint files as Mini-Splatting \cite{fang2024mini} before our “optimizing-sparsifying” process to ensure fair comparisons. \proj~starts “optimizing-sparsifying” at iteration 15K and removes ``zero" Gaussians at iteration 25K. 

\begin{figure}[t!]
    \centering
    \includegraphics[width=\linewidth]{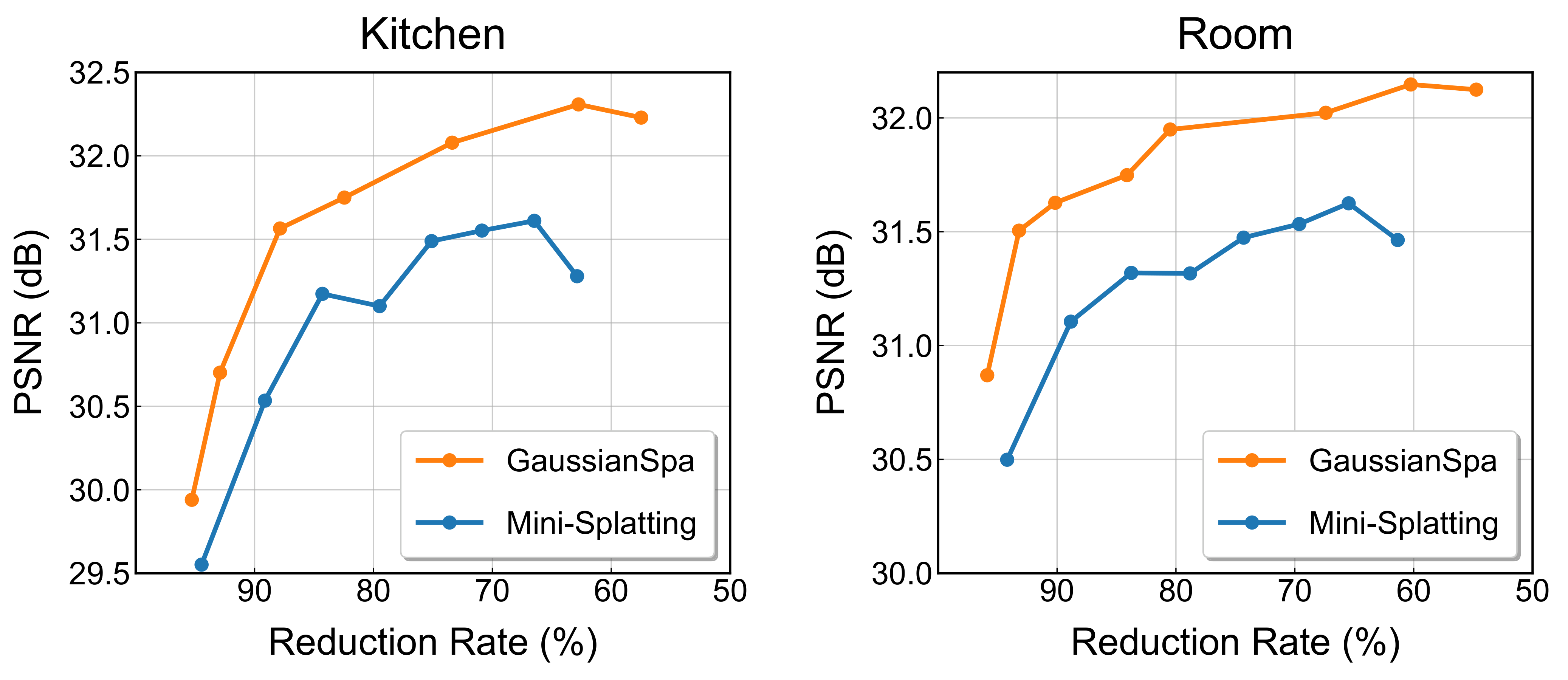}
    \vspace{-8mm}
    \caption{\textbf{Quality-Reduction Rate curves on the (left) Kitchen and (right) Room scenes.} \proj~outperforms Mini-Splatting \cite{fang2024mini} at multiple reduction rates.}
    \vspace{-1.5mm}
    \label{fig:psnr_over_iter}
\end{figure}

\subsection{Quantitative Results}

The quantitative results are summarized in Table \ref{tab:Quantitative_comparision}, compared with various existing approaches. The table shows that our proposed \proj~outperforms the baseline 3DGS across all three metrics with substantially fewer Gaussians. To be specific, after reducing the Gaussian points by 6$\times$, \proj~still improves the PSNR by 0.4 dB over the vanilla 3DGS. Compared to other criterion-based simplification methods, including EAGLES \cite{girish2023eagles}, Mini-Splatting \cite{fang2024mini}, Taming 3DGS \cite{mallick2024taming}, and CompGS \cite{navaneet2024compgs}, as well as the learnable mask-based approaches, CompactGaussia \cite{lee2024compact} and LP-3DGS \cite{zhang2024lp}, our \proj~exhibits a significant improvement under a less number of Gaussians. Particularly, \proj~achieves a PSNR improvement as high as 0.7 dB with even 3$\times$ fewer Gaussians, compared against LP-3DGS \cite{zhang2024lp}. These results quantitatively demonstrate the superiority of our \proj.

We plot quality curves for the Gaussian reduction rate on the Room and Kitchen in Figure \ref{fig:psnr_over_iter} (more results are in Figure \ref{fig:psnrAll} in the Supplementary Material), showing the robustness of \proj~for different numbers of remaining Gaussians. \proj~outperforms the state-of-the-art method, Mini-Splatting \cite{fang2024mini}, by an average of 0.5 dB PSNR improvement across multiple Gaussian reduction rates.

\begin{figure*}[t!]
    \centering
    \includegraphics[width=\linewidth]{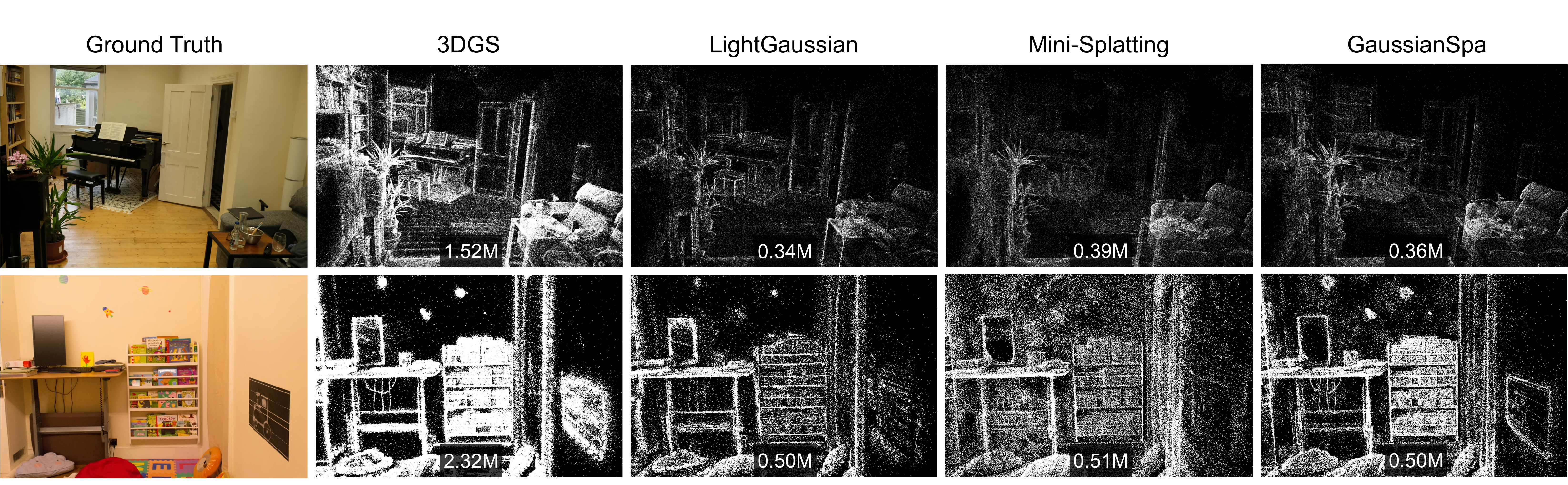}
    \vspace{-8mm}
    \caption{\textbf{Visualized point clouds for the Room and Drjohnson scenes.} With \proj, sparse Gaussians concentrate on high-frequency areas after our ``optimizing-sparsifying"-integrated training process, ensuring the capability to capture detail-rich textures and shapes. On the contrary, the point clouds produced by other approaches cannot outline the contours.}
    \label{fig:visual-results-point-cloud}
\end{figure*}

\subsection{Visual Quality Results}%

Figure \ref{fig:RGBrender4exp} shows the rendered images on the Drjohnson and Counter scenes \cite{barron2022mipnerf360, hedman2018deep} for visual quality evaluation, in comparison to LightGaussian \cite{fan2023lightgaussian}, Mini-Splatting \cite{fang2024mini}, and the vanilla 3DGS. Together with Figure \ref{fig:intro-cover-comp}, these outcomes demonstrate \proj's superior rendering quality on diverse scenes over existing approaches, consistent with the quantitative results in the previous subsection. For instance, as shown in Figure \ref{fig:RGBrender4exp}, \proj~accurately renders the outlet behind the railings for the Playroom scene, while LightGaussian \cite{fan2023lightgaussian} and Mini-Splatting \cite{fang2024mini} lose this object. Furthermore, \proj~fully recovers the original details of the zoom-in object in the Counter scene, while other methods outcome blurry surfaces.

We analyze the proposed \proj~by visualizing the rendered Gaussian ellipsoids and final point clouds in Figure \ref{fig:visual-results-ellipsoids} and Figure \ref{fig:visual-results-point-cloud}, respectively. Those visualized results demonstrate that \proj~can adaptively control the density to match the details when sparsifying the Gaussians, ensuring high-quality sparse 3D representation. From Figure \ref{fig:visual-results-ellipsoids} and Figure \ref{fig:app-visual-results-ellipsoids}, we can observe that \proj~uses fewer but larger Gaussians to represent the blue sky and uses dense Gaussians to render complex carpet textures, providing precise rendering of details in the zoom-in images. In contrast, Mini-Splatting \cite{fang2024mini} cannot fully present the details, leading to blurry rendered textures. Figure \ref{fig:visual-results-point-cloud} further demonstrates the superiority of \proj~in simplification over existing methods. \proj~adaptively reduce the Gaussians in the low-frequency areas and preserve more Gaussians in high-frequency areas that generally need denser Gaussians to synthesize, resulting in sparse Gaussian point clouds that precisely outline the scene contours.

\section{Discussion}
\label{sec:discussion}

\subsection{Generality of \proj}
The proposed \proj~is a general Gaussian simplification framework. Existing importance criteria can be applied to project the auxiliary variable $\bm{z}$ onto the sparse space in the ``sparsifying" step, Eq. \ref{eqn:obj-sparsifying}, obtaining more compact 3DGS models with improved rendering performance. For example, we employ the criterion from LightGaussian \cite{fan2023lightgaussian} and plot the PSNR curves at multiple Gaussian reduction rates in Figure \ref{fig:psnrAll} in the Supplementary Material. It is observed that \proj~consistently outperforms LightGaussian \cite{fan2023lightgaussian} with an average of 0.4 dB improvement for the Playroom and Kitchen scenes.

\subsection{Storage Compression}

While \proj~focuses on simplifying and optimizing the number of Gaussians for a sparse 3DGS representation, existing compression techniques such as SH optimization and vector quantization can be applied as add-ons to the simplified 3DGS model. Table \ref{tab:storageCompare} (in Supplementary Material) shows the final storage cost for the Mip-NeRF 360 dataset by incorporating the SH distillation and vector quantization methods proposed by LightGaussian \cite{fan2023lightgaussian}. This comparison demonstrates that \proj~can achieve the lowest storage consumption while maintaining superior rendering quality over existing hybrid compression methods, such as EfficientGS \cite{liu2024efficientgs} and LightGaussian \cite{fan2023lightgaussian}.

\subsection{Relation to Other Optimization Algorithms}
The ``optimizing-sparsifying" solution in our proposed \proj~falls under optimization methods that project a target variable onto a constrained space,  specifically the $\ell_p$ ball (where $p = 0$ in our case). This approach is similar to foundational optimization algorithms in compressed sensing \cite{donoho2006compressed} and sparse learning \cite{bruckstein2009sparse}, such as Ridge regression \cite{hoerl1970ridge} with the $\ell_2$ regularization and Lasso regression \cite{tibshirani1996regression} with the $\ell_1$ regularization. The regularization acts as a penalty, combined with quadratic loss, reformulating the problem as a convex optimization task.

However, genuine sparsity requires $\ell_0$ regularization, which leads to a nonconvex, NP-hard problem \cite{oktar2018review, davis1994adaptive, natarajan1995sparse, tillmann2014computational}. Lasso (where $p = 1$) provides the tightest convex approximation but sacrifices exact sparsity \cite{chen2001atomic}. Greedy algorithms like Matching Pursuit \cite{mallat1993matching, pati1993orthogonal} offer another approach, though they lack guaranteed global optima.

Iterative thresholding (IT) \cite{elad2007wide, daubechies2004iterative, eldar2012compressed} offers a different solution, using matrix multiplications and scalar shrinkage steps to yield a simple yet effective structure. Furthermore, a closed-form solution for the global minimizer can be obtained in cases with a unitary matrix \cite{elad2007wide}, even for non-convex scenarios. Inspired by this, we introduce a duplicated variable (i.e., $\bm{z}$) and project it onto the $\ell_0$ ball, optimizing over two variables alternately, similar to the alternating direction method of multipliers \cite{boyd2011distributed}.  Leveraging the unitary case's simplicity, the ``sparsifying" step aligns with this scenario and thus admits an explicit solution. Thus, our algorithm benefits from $\ell_0$-induced sparsity while achieving fast convergence guided by the explicit solution.

\section{Conclusion}
In this paper, we present an optimization-based simplification framework, \proj, for compact and high-quality 3DGS. \proj~formulates the simplification objective as a constrained optimization problem with a sparsity constraint on the Gaussian opacities. Then, we propose an ``optimizing-sparsifying" solution to efficiently solve the formulated problem during the 3DGS training process. Our comprehensive evaluations on multiple datasets with quantitative results and qualitative analyses demonstrate the superiority of \proj~in rendering quality with fewer Gaussians compared to the state of the arts.

\clearpage

{
    \small
    \bibliographystyle{ieeenat_fullname}
    \bibliography{main}
}
\clearpage
\setcounter{page}{1}
\maketitlesupplementary

\section{Additional Results}

\subsection{Additional Quantitative Results}
\begin{figure*}[t!]
    \centering
    \includegraphics[width=\linewidth]{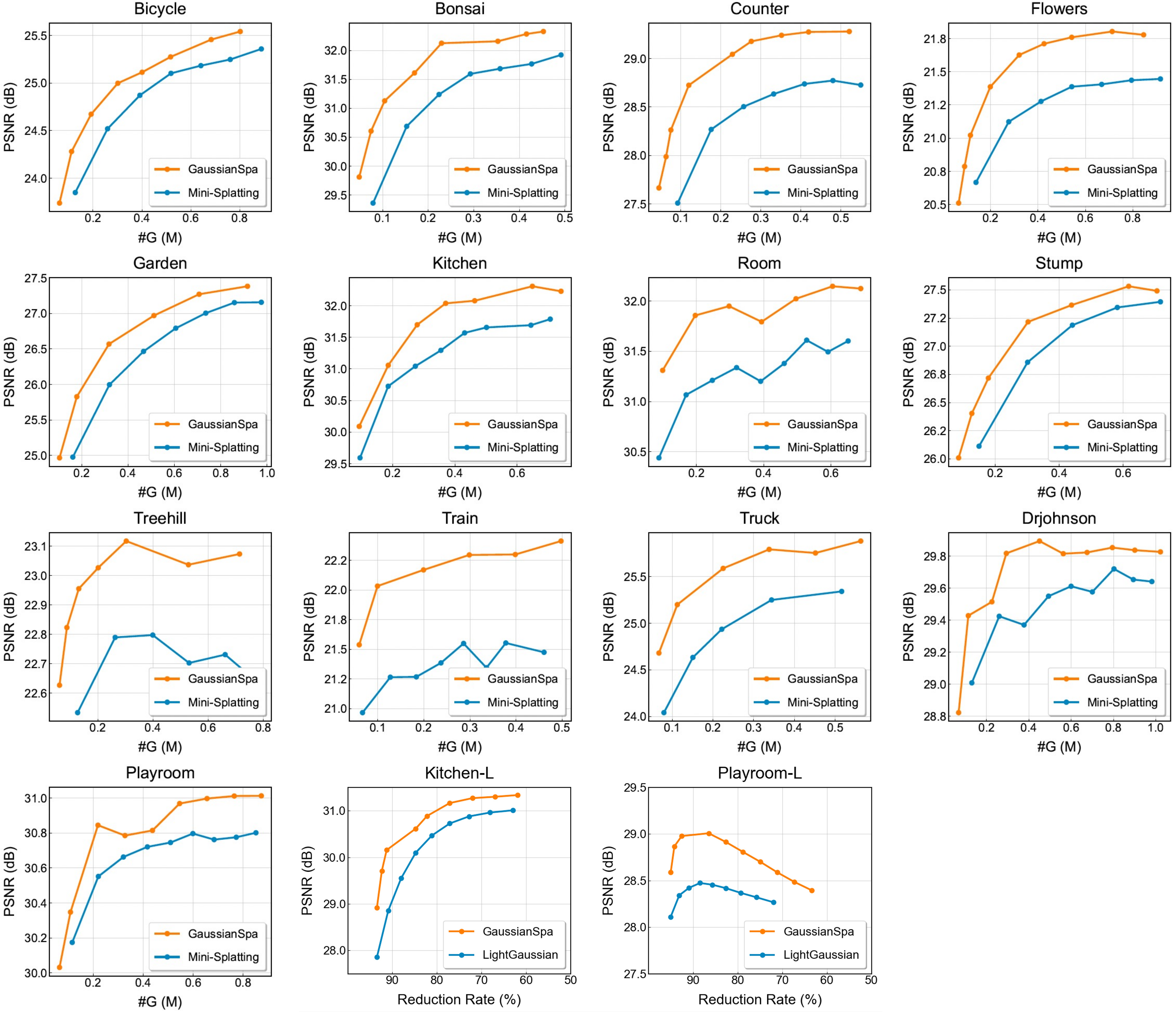}
    \vspace{-6mm}
    \caption{The first 13 sub-figures: Quality-\#G (the number of Gaussians in millions) curves comparing \proj~with Mini-Splatting \cite{fang2024mini} on multiple scenes. Our \proj~consistently outperforms Mini-Splatting \cite{fang2024mini} with the same \#G. The last two sub-figures: Quality-Reduction Rate curves comparing \proj~with LightGaussian \cite{fan2023lightgaussian} on the Kitchen and Playroom scenes.}
    \label{fig:psnrAll}
\end{figure*}

We summarize additional quantitative results on the Mip-NeRF 360, Tanks\&Temples, and Deep Blending datasets in Table \ref{tab:MipAverAppendix}, Table \ref{tab:TanksTemplesAverAppendix}, and Table \ref{tab:DeepBlendingAverAppendix}, respectively. We plot additional PSNR-\#Gaussians curves on diverse Scenes in Figure \ref{fig:psnrAll}, in comparison with Mini-Splatting \cite{fang2024mini} and LightGaussian \cite{fan2023lightgaussian}. It can be observed that with the same number of Gaussians, our \proj~shows superior rendering outcomes.

\subsection{Additional Visual Quality Results}
\begin{figure*}[t!]
    \centering
    \includegraphics[width=\linewidth]{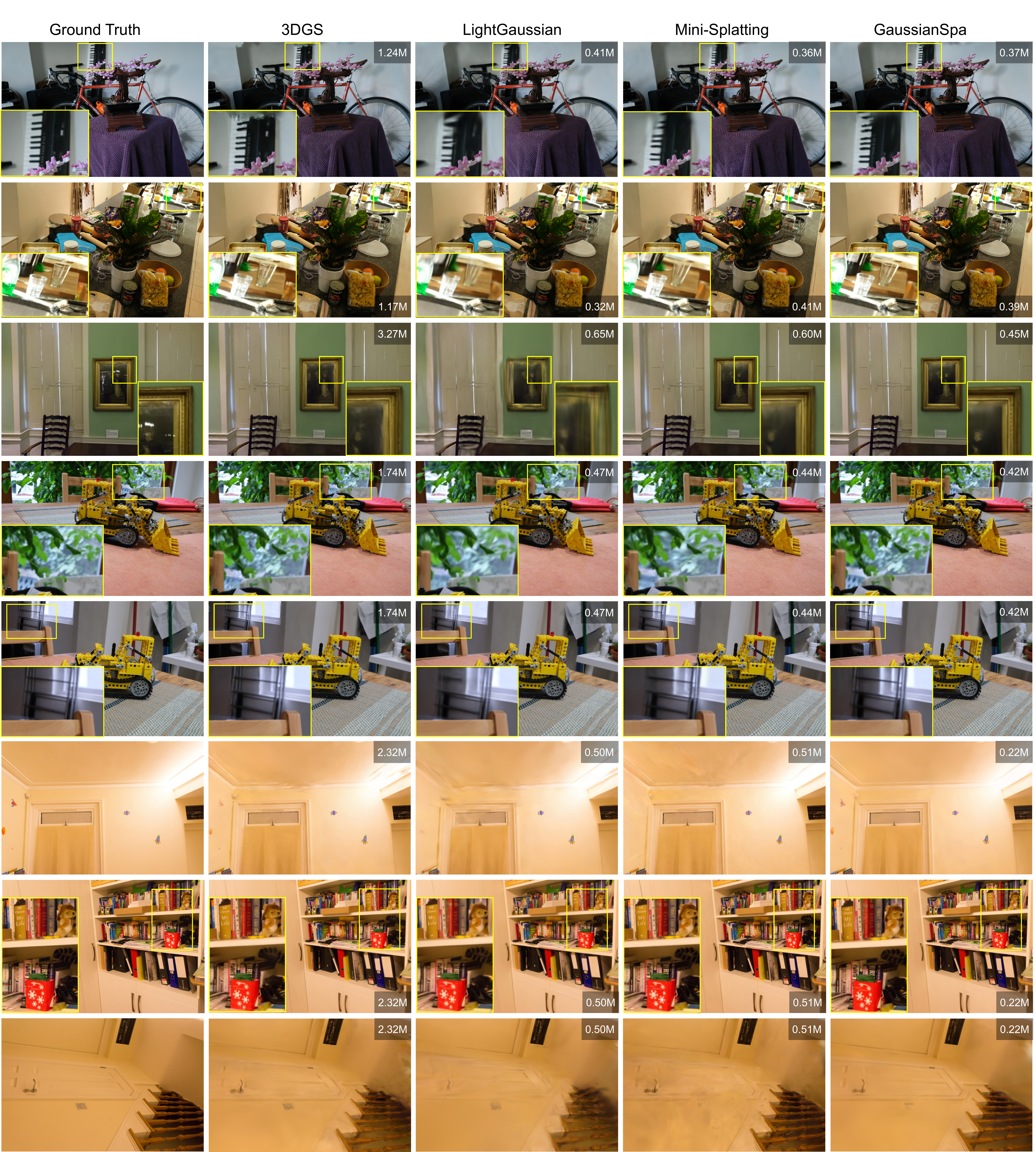}
    \caption{Additional visual comparison on more scenes. The numbers of remaining Gaussians in millions are displayed.}
    \label{fig:app-RGB-Mip}
\end{figure*}

\begin{figure*}[t!]
    \centering
    \includegraphics[width=\linewidth]{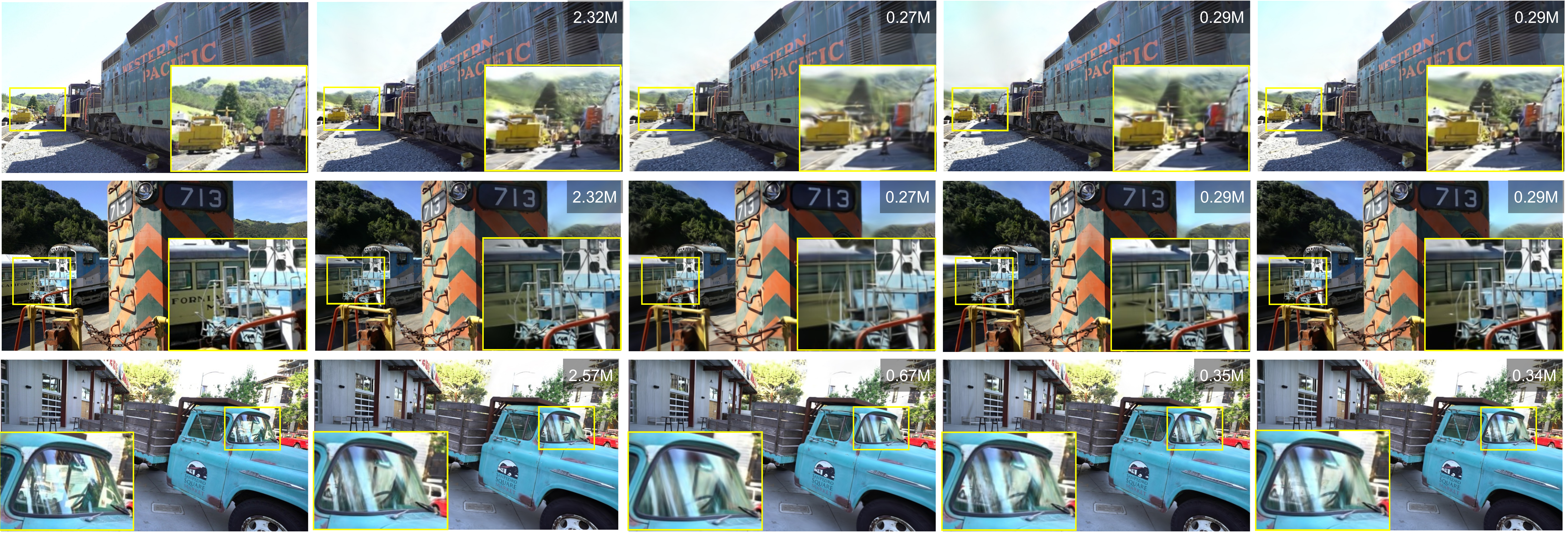}
    \vspace{-6mm}
    \caption{(Continue) Additional visual comparison on more scenes. The numbers of remaining Gaussians in millions are displayed.}
    \label{fig:app-RGB-Truck}
    \vspace{+1.5mm}
\end{figure*}

\begin{figure*}[t!]
    \centering
    \includegraphics[width=\linewidth]{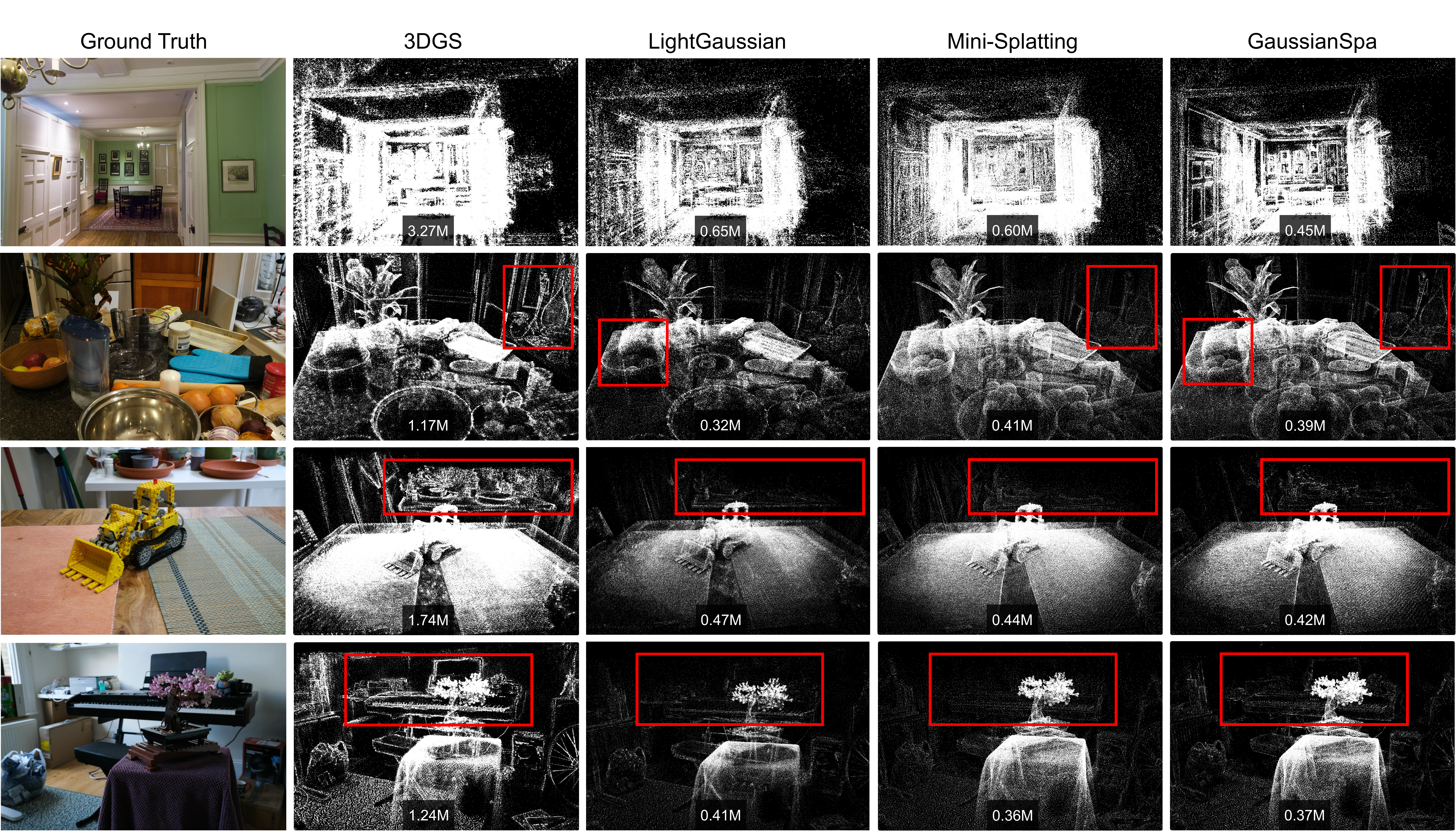}
    \vspace{-6mm}
    \caption{Additional visualized point clouds. The numbers of remaining Gaussians in millions are displayed.}
    \label{fig:additional_distribution_compare}
    \vspace{-3mm}
\end{figure*}

Figure \ref{fig:app-RGB-Mip} and Figure \ref{fig:app-RGB-Truck} provide additional rendered images comparing \proj~with LightGaussian \cite{fan2023lightgaussian}, Mini-Splatting \cite{fang2024mini}, and original 3DGS on various scenes. Those additional results demonstrate our \proj~achieves stronger representational power for background and detail-rich areas such as walls, carpets, and ladders, showcasing superior rendering qualities. Furthermore, Figure \ref{fig:app-visual-results-ellipsoids} and Figure \ref{fig:additional_distribution_compare} offer additional rendered Gaussian ellipsoids and point cloud views, respectively. These results further illustrate that \proj~creates a high-quality sparse 3D representation that adaptively uses more Gaussians to represent high-frequency areas.

\section{Additional Discussion}

\begin{table}[b!]
  \centering
    \resizebox{\linewidth}{!}{%
    \begin{tabular}{lcccc}
    \toprule
    Method  & PSNR↑  & SSIM↑  & LPIPS↓ & Storage↓ \\
    \midrule
    EfficientGS \cite{liu2024efficientgs} & 27.38 & 0.817 & 0.216 & 98 MB \\
    LightGaussian \cite{fan2023lightgaussian} & 27.28 & 0.805 & 0.243 & 42 MB \\
    \rowcolor[rgb]{ .902,  .902,  .902} \textbf{GaussianSpa} & \textbf{27.85} & \textbf{0.825} & \textbf{0.214} & \textbf{25 MB} \\
    \bottomrule
    \end{tabular}%
    }
    \vspace{-3mm}
    \caption{\textbf{Storage comparison evaluated on the Mip-NeRF 360 dataset.} \proj’s storage cost is reported based on the add-on compression methods (i.e., SH distillation and vector quantization) from LightGaussian \cite{fan2023lightgaussian}. }
    \vspace{-1mm}
  \label{tab:storageCompare}%
\end{table}%

\begin{figure}[t!]
    \centering
    \begin{subfigure}[b]{0.49\linewidth}
        \includegraphics[width=\linewidth]{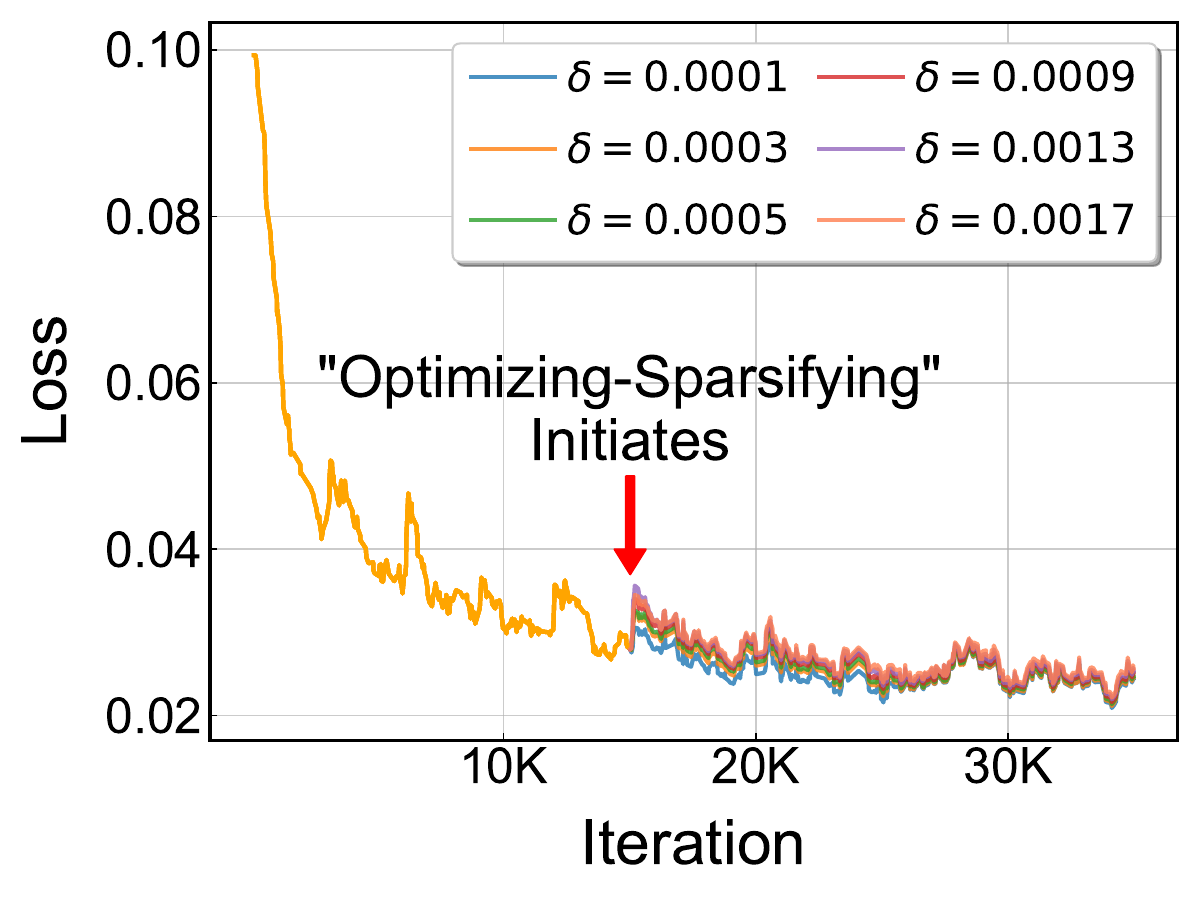}
    \end{subfigure}
    \hfill
    \begin{subfigure}[b]{0.49\linewidth}
        \includegraphics[width=\linewidth]{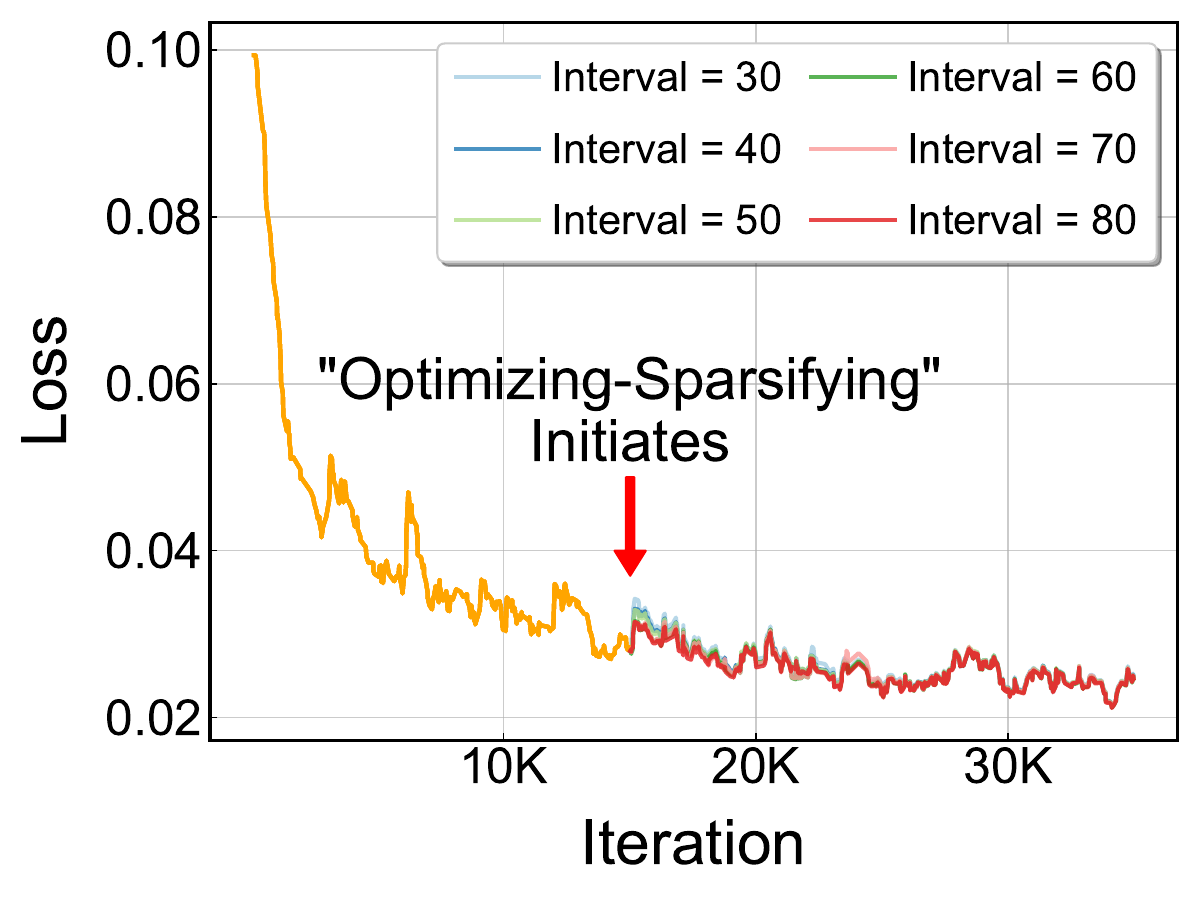}
    \end{subfigure}
    \vspace{-3mm}
    \caption{\textbf{Loss curves with multiple (left) $\delta$ and (right) interval settings.} The curves show that \proj~exhibits a good convergence behavior in the ``optimizing-sparsifying"-integrated training process. }
    \vspace{-5mm}
    \label{fig:rholoss}
\end{figure}

\textbf{Convergence Analysis.}
We analyze the convergence behavior of \proj~by examining the effects of hyper-parameters such as $\delta$, which controls the sparsity strength in Eq. \ref{eqn:obj-lagrangian}, and the interval at which the ``sparsifying" step is performed, as described in Section \ref{sec:method}. 
Figure \ref{fig:rholoss} shows loss curves for various $\delta$ and interval values, indicating similar convergence rates during our ``optimizing-sparsifying"-integrated training process. After removing ``zero" Gaussians at iteration 25K, the loss curve continues to converge consistently, confirming \proj's feasibility.

\begin{table}[t!]
  \centering
    \resizebox{\linewidth}{!}{%
    \begin{tabular}{cccccc}
    \toprule
    Scene & Method & PSNR↑  & SSIM↑  & LPIPS↓ & \# G (M)↓ \\
    \midrule
    \multirow{3}{*}{Bicycle} & 3DGS  & 25.13 & 0.750 & \textbf{0.240} & 5.310 \\
          & Mini-Splatting  & 25.21 & 0.760 & 0.247 & \textbf{0.646} \\
          & GaussianSpa & \textbf{25.44} & \textbf{0.769} & 0.246 & 0.656 \\
    \midrule
    \multirow{3}{*}{Bonsai} & 3DGS  & 32.19 & \textbf{0.950} & 0.180 & 1.250 \\
          & Mini-Splatting  & 31.73 & 0.945 & 0.180 & \textbf{0.360} \\
          & GaussianSpa & \textbf{32.40} & 0.947 & \textbf{0.174} & 0.372 \\
    \midrule
    \multirow{3}{*}{Counter} & 3DGS  & 29.11 & 0.910 & 0.180 & 1.170 \\
          & Mini-Splatting  & 28.53 & 0.911 & 0.184 & 0.408 \\
          & GaussianSpa & \textbf{29.23} & \textbf{0.919} & \textbf{0.176} & \textbf{0.392} \\
    \midrule
    \multirow{3}{*}{Flowers} & 3DGS  & 21.37 & 0.590 & 0.360 & 3.470 \\
          & Mini-Splatting  & 21.42 & \textbf{0.616} & 0.336 & \textbf{0.670} \\
          & GaussianSpa & \textbf{21.75} & 0.610 & \textbf{0.329} & 0.674 \\
    \midrule
    \multirow{3}{*}{Garden} & 3DGS  & \textbf{27.32} & \textbf{0.860} & \textbf{0.120} & 5.690 \\
          & Mini-Splatting  & 26.99 & 0.842 & 0.156 & 0.738 \\
          & GaussianSpa & 27.26 & 0.848 & 0.151 & \textbf{0.728} \\
    \midrule
    \multirow{3}{*}{Kitchen} & 3DGS  & 31.53 & 0.930 & 0.120 & 1.770 \\
          & Mini-Splatting & 31.24 & 0.929 & 0.122 & 0.438 \\
          & GaussianSpa & \textbf{32.03} & \textbf{0.934} & \textbf{0.117} & \textbf{0.423} \\
    \midrule
    \multirow{3}{*}{Room} & 3DGS  & 31.59 & 0.920 & 0.200 & 1.500 \\
          & Mini-Splatting & 31.44 & 0.929 & 0.193 & 0.394 \\
          & GaussianSpa & \textbf{32.04} & \textbf{0.933} & \textbf{0.188} & \textbf{0.355} \\
    \midrule
    \multirow{3}{*}{Stump} & 3DGS  & 26.73 & 0.770 & 0.240 & 4.420 \\
          & Mini-Splatting  & 27.35 & 0.803 & 0.219 & 0.717 \\
          & GaussianSpa & \textbf{27.56} & \textbf{0.808} & \textbf{0.218} & \textbf{0.690} \\
    \midrule
    \multirow{3}{*}{Treehill} & 3DGS  & 22.61 & 0.640 & 0.350 & 3.420 \\
          & Mini-Splatting  & 22.69 & 0.652 & 0.332 & 0.663 \\
          & GaussianSpa & \textbf{22.94} & \textbf{0.660} & \textbf{0.329} & \textbf{0.637} \\
    \midrule
    \multirow{3}{*}{Average} & 3DGS  & 27.45 & 0.810 & 0.220 & 3.110 \\
          & Mini-Splatting  & 27.40 & 0.821 & 0.219 & 0.559 \\
          & GaussianSpa & \textbf{27.85} & \textbf{0.825} & \textbf{0.214} & \textbf{0.547} \\
    \bottomrule
    \end{tabular}%
    }
    \vspace{-3mm}
    \caption{MiP-NeRF360 per scene results. 3DGS results are reported from \cite{girish2023eagles}. Mini-Splatting \cite{fang2024mini} results are replicated using official code.}
  \label{tab:MipAverAppendix}%
\end{table}%

\begin{figure*}[t!]
    \centering
    \includegraphics[width=\linewidth]{figs/ellipsoid_drjohnson.pdf}
    \vspace{-6mm}
    \caption{Additional visualized Gaussian ellipsoids. The numbers of remaining Gaussians in millions are displayed.}
    \label{fig:app-visual-results-ellipsoids}
    \vspace{-3mm}
\end{figure*}

\begin{table*}[t!]
    \centering
    \begin{minipage}[b]{0.49\linewidth}
    \resizebox{1\linewidth}{!}{%
        \begin{tabular}{cccccc}
        \toprule
        Scene & Method & PSNR↑  & SSIM↑  & LPIPS↓ & \# G (M)↓ \\
        \midrule
        \multirow{3}{*}{Train} & 3DGS  & 21.94 & 0.810 & \textbf{0.200} & 1.110 \\
              & Mini-Splatting  & 21.78 & 0.805 & 0.231 & 0.287 \\
              & GaussianSpa & \textbf{22.17} & \textbf{0.815} & 0.228 & \textbf{0.199} \\
        \midrule
        \multirow{3}{*}{Truck} & 3DGS  & 25.31 & 0.880 & 0.150 & 2.540 \\
              & Mini-Splatting  & 25.13 & 0.878 & 0.141 & 0.352 \\
              & GaussianSpa & \textbf{25.79} & \textbf{0.888} & \textbf{0.132} & \textbf{0.338} \\
        \midrule
        \multirow{3}{*}{Average} & 3DGS  & 23.63 & 0.850 & \textbf{0.180} & 1.830 \\
              & Mini-Splatting  & 23.45 & 0.841 & 0.186 & 0.319 \\
              & GaussianSpa & \textbf{23.98} & \textbf{0.852} & \textbf{0.180} & \textbf{0.269} \\
        \bottomrule
        \end{tabular}%
        }
        \caption{Tanks\&Temples per scene results.}
        \label{tab:TanksTemplesAverAppendix}
    \end{minipage}%
    \hfill
    \begin{minipage}[b]{0.49\linewidth}
    \resizebox{1\linewidth}{!}{%
        \begin{tabular}{cccccc}
        \toprule
        Scene & Method & PSNR↑  & SSIM↑  & LPIPS↓ & \# G (M)↓ \\
        \midrule
        \multirow{4}{*}{Drjohnson} & 3DGS  & 28.77 & 0.900 & 0.250 & 3.260 \\
              & Mini-Splatting & 29.37 & 0.904 & 0.261 & 0.377 \\
              & GaussianSpa & \textbf{29.89} & \textbf{0.913} & \textbf{0.243} & 0.450 \\
              & GaussianSpa & 29.82 & 0.909 & 0.254 & \textbf{0.293} \\
        \midrule
        \multirow{4}{*}{Playroom} & 3DGS  & 30.07 & 0.900 & 0.250 & 2.290 \\
              & Mini-Splatting  & 30.72 & 0.914 & \textbf{0.248} & 0.417 \\
              & GaussianSpa  & \textbf{30.84} & \textbf{0.916} & 0.254 & \textbf{0.219} \\
              & GaussianSpa  & \textbf{30.84} & \textbf{0.916} & 0.254 & \textbf{0.219} \\
        \midrule
        \multirow{4}{*}{Average} & 3DGS  & 29.42 & 0.900 & 0.250 & 2.780 \\
              & Mini-Splatting  & 30.05 & 0.909 & 0.254 & 0.397 \\
              & GaussianSpa & \textbf{30.37} & \textbf{0.914} & \textbf{0.249} & 0.335 \\
              & GaussianSpa & 30.33 & 0.912 & 0.254 & \textbf{0.256} \\
        \bottomrule
        \end{tabular}%
        }
        \caption{Deep Blending per scene results.}
        \label{tab:DeepBlendingAverAppendix}
    \end{minipage}
\end{table*}

\end{document}